\documentclass[letterpaper, 10 pt, conference]{ieeeconf}
\usepackage{graphicx}
\usepackage{cite}
\usepackage{hyperref}
\usepackage[font=small]{caption}
\usepackage{subcaption}
\usepackage{amsmath}
\usepackage{bm}
\usepackage{amsfonts}
\usepackage{array}

\IEEEoverridecommandlockouts
\begin{document}
\title{\huge Point Cloud Compression with Bits-back Coding}
\author{Nguyen Quang Hieu, Minh Nguyen, Dinh Thai Hoang, Diep N. Nguyen, and Eryk Dutkiewicz 
\thanks{N. Q. Hieu, D. T. Hoang, D. N. Nguyen, and E. Dutkiewicz are with the School of Electrical and Data Engineering, University of Technology Sydney, Australia. }
\thanks{M. Nguyen is with Fraunhofer FOKUS, Germany.}}
\maketitle
\begin{abstract}
This paper introduces a novel lossless compression method for compressing geometric attributes of point cloud data with bits-back coding. Our method specializes in using a deep learning-based probabilistic model to estimate the Shannon's entropy of the point cloud information, i.e., geometric attributes of the 3D floating points. Once the entropy of the point cloud dataset is estimated with a convolutional variational autoencoder (CVAE), we use the learned CVAE model to compress the geometric attributes of the point clouds with the bits-back coding technique. The novelty of our method with bits-back coding specializes in utilizing the learned latent variable model of the CVAE to compress the point cloud data. By using bits-back coding, we can capture the potential correlation between the data points, such as similar spatial features like shapes and scattering regions, into the lower-dimensional latent space to further reduce the compression ratio. The main insight of our method is that we can achieve a competitive compression ratio as conventional deep learning-based approaches, while significantly reducing the overhead cost of storage and/or communicating the compression codec, making our approach more applicable in practical scenarios. 
Throughout comprehensive evaluations, we found that the cost for the overhead is significantly small, compared to the reduction of the compression ratio when compressing large point cloud datasets. Experiment results show that our proposed approach can achieve a compression ratio of 1.56 bit-per-point on average, which is significantly lower than the baseline approach such as Google's Draco with a compression ratio of 1.83 bit-per-point.
\end{abstract}


\section{Introduction}
Point cloud offers a flexible and visually rich representation of 3D data, with the ability to capture detailed spatial information. This makes point cloud invaluable in a wide range of applications, including automotive LiDAR for autonomous driving, 3D scene understanding in robotics, and immersive experiences in virtual reality and augmented reality. Despite the flexibility and usefulness of the point cloud, the unstructured format of the point cloud poses significant challenges in data storage and often requires efficient data compression techniques \cite{cao20193d}. For example, a single sweep of LiDAR sensors in autonomous driving can produce 100,000 points, resulting in 84 billion points per day \cite{huang2020octsqueeze}. Moreover, a single point in the point cloud may require 32 or 64 bits to represent the geometric attributes, i.e., $(x, y, z)$ position in a 3D coordinate, let alone the color attributes, resulting in a massive amount of data storage.
However, the unordered nature of point clouds presents a significant challenge for efficient compression. Consequently, point cloud compression has emerged as a critical area of research, focusing on transforming raw point cloud data into a more structured format that can be more efficiently compressed \cite{cao20193d}.

\begin{figure}[t]
\centering
\includegraphics[width=0.95\linewidth]{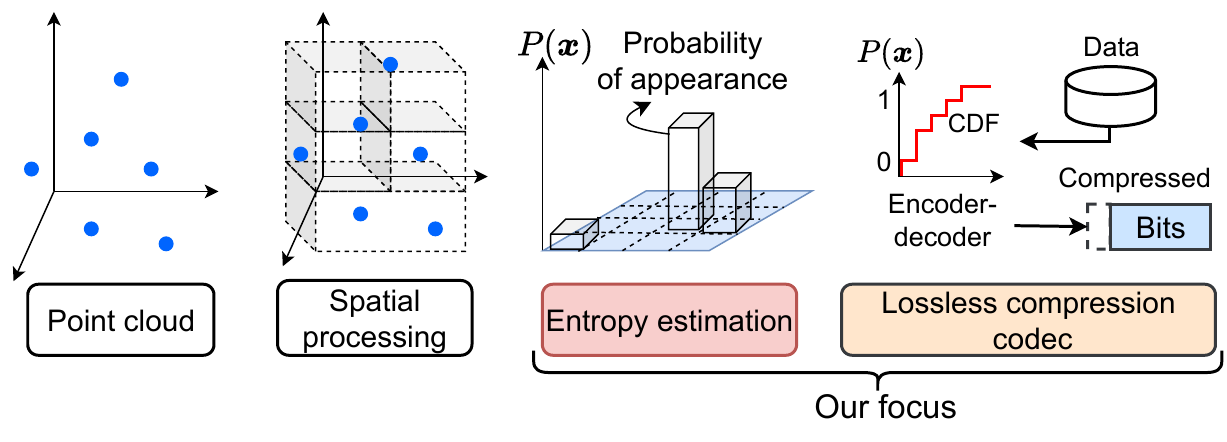}
\caption{Overview of a point cloud compression pipeline. Decompression can be done in a reversed order.}
\label{fig:pcc-overview}
\end{figure}

A typical point cloud compression process can be illustrated in Fig.~\ref{fig:pcc-overview}, beginning with ``spatial processing", which converts raw point cloud data into ordered formats such as tree-based structures or voxel grids. The spatial processing step is often a lossy process and is critical as it facilitates the subsequent transformation of the data into a binary sequence. Once organized in the tree-based or voxel format, the data can be processed by a lossless compression codec, which involves an important step of entropy estimation. The entropy estimation step is critical to quantify the uncertainty or randomness in the data, typically using statistical tools like histograms or machine learning models to approximate Shannon's entropy \cite{huang2020octsqueeze, nguyen2021learning, wang2021lossy, he2022density}. The lossless compression codec then utilizes the estimated entropy, typically conditioned on the marginal probability of the data, to produce the final binary sequence, as depicted in Fig.~\ref{fig:pcc-overview}.
The most widely used lossless compression codecs are arithmetic coding \cite{witten1987arithmetic} and the recently developed asymmetric numeral systems (ANS) \cite{duda2013asymmetric}, thanks to their implementation simplicity and low compression overhead. The main idea of these codecs is to build the cumulative distribution functions (CDFs) and the inverse CDFs from the learned marginal distribution of data, denoted as $P(\bm{x})$ in Fig.~\ref{fig:pcc-overview}. Once the learned distribution functions are shared between the encoder (sender) and decoder (receiver), data samples are mapped into a binary sequence by transforming the value $P(\bm{x})$ into an interval between 0 and 1 \cite{duda2013asymmetric, mackay2003information}.

Accurate entropy estimation is crucial, as it directly impacts the efficiency of lossless compression codecs like arithmetic coding and ANS. While considerable research works have focused on enhancing entropy estimation through advanced deep learning models, e.g., \cite{nguyen2021learning, wang2021lossy, he2022density} and references therein, the resulting reduction in compression ratios (defined as the number of bits required to encode a single geometric attribute) has often been modest, underscoring the need for novel methods. A key limitation of current deep learning-based entropy estimation models for point clouds is the assumption that the decoder has access to the learned marginal probability of the data, denoted as $P(\bm{x})$. In many practical scenarios, this assumption is unrealistic, as the encoder and decoder would need to store or communicate the marginal probability. The cost of doing so could exceed the size of the compressed data itself, particularly when the point cloud input scales to hundreds of thousands of points. This will be evaluated in Section \ref{sec:performance-evaluation}, when we investigate the overhead associated with conventional approaches, and demonstrate how to reduce this cost with bits-back coding.

In this work, we propose a novel technique for compressing point clouds based on bits-back coding \cite{frey1996free, townsend2018practical, Townsend2020HiLLoC}. Our approach leverages a latent variable model, where both the encoder and decoder employ a prior distribution over the model parameters to achieve efficient data compression. The core principle of bits-back coding is to encode and decode data by utilizing the prior, posterior, and likelihood from a trained variational autoencoder (VAE) \cite{townsend2018practical, Townsend2020HiLLoC}.
This approach is advantageous for compressing large point cloud sequences, which often exhibit significant correlation between consecutive point clouds, yet do not require access to the marginal probability of the data - an element that can introduce considerable overhead.
As a result, our method results in low overhead costs in terms of codec communication and installation, making it a practical solution for real-world applications requiring efficient and scalable point cloud compression. In summary, our main contributions are as follows:

\begin{itemize}
\item We first develop a simplistic variational autoencoder with 3D convolutional filters to estimate Shannon's entropy of the voxelized point clouds. The main aim of using 3D convolutional filters is to subsequently reduce the input size of the voxel grids and efficiently project the 3D voxel grids into a lower dimensional representation. At its core, our CVAE model has a latent space with Gaussian prior, enabling the model to learn the correlation between point clouds in large datasets and later use of bits-back coding in the inference step.  
\item We then develop a novel compression method based on bits-back coding to compress a batch of point clouds. The main idea of using bits-back coding is to reduce the compression ratio when the size of the point clouds increases. In return, bits-back coding pays an overhead cost of inserting latent variables into the compression process. However, we provide empirical evaluations to show that this overhead cost is trivial compared to the compression gain of the bits-back coding method, as the proposed method outperforms an industrial standard point cloud compression library Draco developed by Google \cite{galligan2018google}.
\end{itemize}

\section{System Model and Proposed Methodology}
\subsection{Spatial Processing}
\label{subsec:spatial-processing}
Recall that the main building blocks of a point cloud compression pipeline is illustrated in Fig.~\ref{fig:pcc-overview} and our main focus in this work is the entropy estimation step and lossless compression step. For the spatial processing step, we employ a simple voxelization processing, in which the entire point cloud is divided into equal-sized 3D cubes called ``voxels". A voxel can be considered as a counterpart of ``pixel" in 2D image/video processing. At the end of the spatial processing step, a raw point cloud is transformed into  $M = 2^{d} \times 2^{d} \times 2^{d}$ voxels in which $d$ is the axis-wise depth value of the processed point cloud. For example, with $d=7$, we use $7$ bits per dimension to represent the $2^7 = 128$ positions of the points in a 3D coordinate. The resulting $M$ voxels now are well structured and can be used for the entropy estimation step. Note that further complicated spatial processing, such as density embedding \cite{he2022density} and building Octrees \cite{huang2020octsqueeze} or KD-trees \cite{galligan2018google}, can be applied for more efficient data presentation but it's not our main focus of this work as we will later show that with simplistic voxelization step described above, we can still outperform other tree-based approaches in terms of compression ratio.

\subsection{Entropy Estimation with CVAE Model}
\label{subsec:entropy-estimation}
\begin{figure}[t]
\centering
\includegraphics[width=0.95\linewidth]{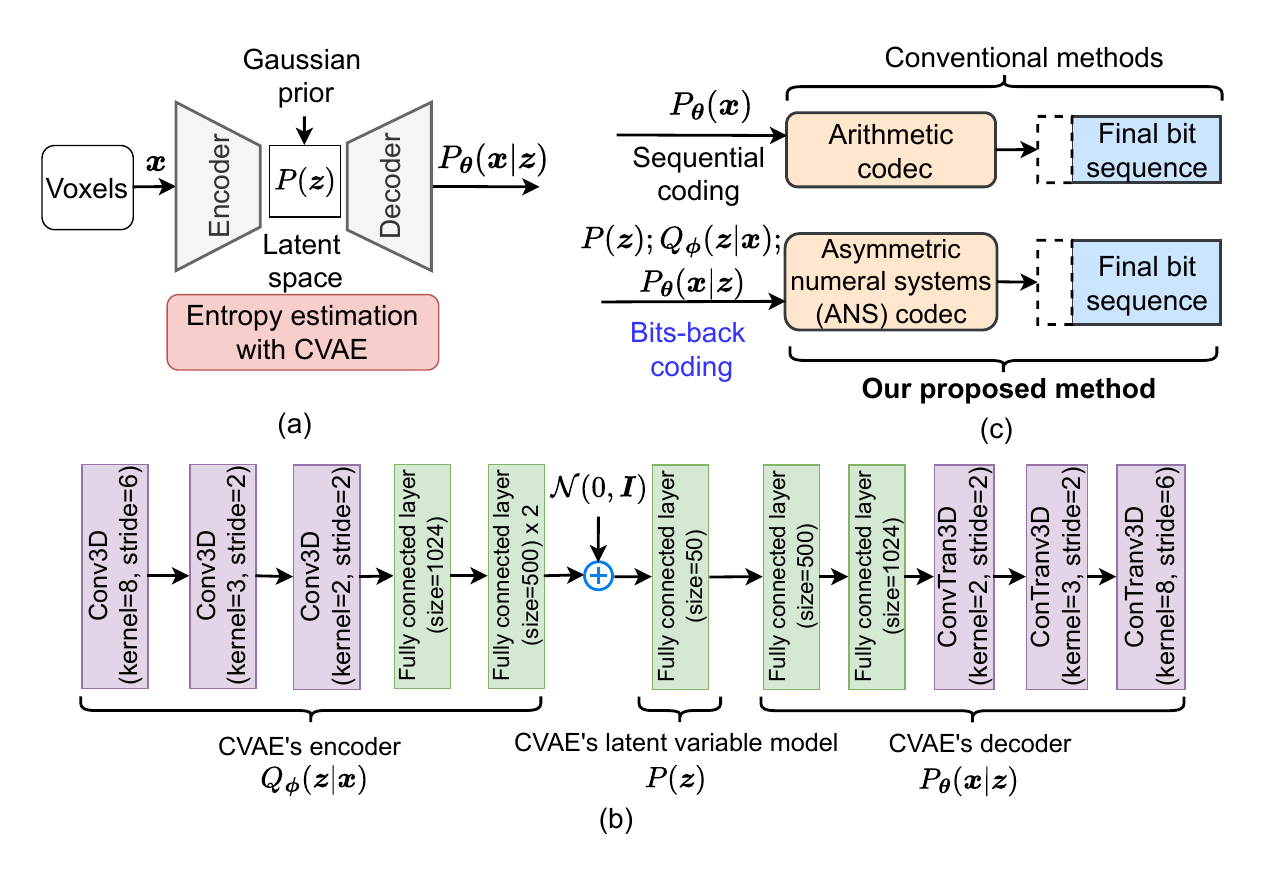}
\caption{(a) Our entropy estimation approach with the proposed Convolutional Variational Autoencoder (CVAE), (b) detailed architecture of the proposed CVAE, and (c) our bits-back coding approach.}
\label{fig:proposed-method}
\end{figure}

In Fig.~\ref{fig:proposed-method}, we illustrate our approach to compress the processed point clouds, i.e., voxels, with bits-back coding. As illustrated in Fig.~\ref{fig:proposed-method}(a), we develop our entropy estimation model for $M$ voxels based on a convolutional variational autoencoder (CVAE). Note that the ``CVAE's encoder" and ``CVAE's decoder" terms in Fig.~\ref{fig:proposed-method}  should not be confused with the encoder-decoder terms used for the arithmetic coding and ANS codecs. We will use the terms ``CVAE's encoder" and ``CVAE's decoder" to explicitly refer to the components of the CVAE model, thus avoiding potential confusion with general encoder-decoder terms used for lossless compression codecs like arithmetic codec and ANS.
It is also noted that, in this work, we only consider compressing the geometric attributes of the point clouds, i.e., the positional values of the floating points in the 3D coordinate, and discard the color information of the points to focus extensively on demonstrating the bits-back coding capabilities. 
We formulate the lossless compression problem of $M$ voxels as follows. 

Given an alphabet $\mathcal{A}$ (i.e., codebook) of voxels having value either 1 or 0, i.e., $\mathcal{A} = \{0, 1\}$, a processed point cloud having $M$  voxels can be represented as a symbol $\bm{x} = [x_1, x_2, \ldots, x_{M}]$ (see footnote\footnote{Note that the symbol $\bm{x}$ can be viewed as a flattened array of voxels for illustration purpose only. In actual implementation, a 3-dimensional array can be utilized for preserving the spatial structure of the point cloud data.} below), in which each element $x_m$ ($1 \leq m \leq M$) receives value either 0 or 1, i.e., $x_m \in \{0, 1\}$. In the context of entropy estimation, let's define the real-valued probability of the variable $x_m$ receiving value $i$ ($i=0, 1$) as $P(x_m=i)$, where $\sum_{i=0}^{1}P(x_m=i) = 1$. We have the Shannon information content of an outcome $x_m$ is $h(x_m = i) = \log \frac{1}{p_i}$,
and the binary entropy function of $x_m$ is
\begin{equation}
H(x_m) =  \sum_{i=0}^{1}P(x_m=i) \log \frac{1}{P(x_m=i)}.
\label{eq:entropy-definition}
\end{equation}
Note that from the above equation and hereafter we use $\log$ to denote the base two logarithm (i.e., usually denoted as $\log_2$) for notational simplicity. For compression of $M$ variable $x_m$, where each $x_m$ is an element of the voxelized point cloud $\bm{x}$, the entropy of $\bm{x}$ can be calculated by \cite[Equation 8.1]{mackay2003information}:
\begin{equation}
H(\bm{x}) = \sum_{x_m \in \mathcal{A}} P(x_1, x_2, \ldots, x_M) \log \frac{1}{P(x_1, x_2, \ldots, x_M)} \\
\label{eq:marginal-entropy}
\end{equation}

In practice, one often requires probabilistic models to estimate the exact marginal probability $P(\bm{x}) = P(x_1, x_2, \ldots, x_M)$ of the vector $\bm{x} = [x_1, x_2, \ldots, x_{M}]$. Let's denote $P_{\bm{\theta}}(x_m =i|x_1, \ldots, x_{m-1})$ being the estimated conditional probability with parameters $\bm{\theta}$ to the true conditional probability $P(x_m = i|x_1, \ldots, x_{m-1})$. Once the conditional probabilities are estimated, the approximate marginal probability $P_{\bm{\theta}}(\bm{x})$ can be calculated using the chain rule, i.e.,
\begin{equation}
P_{\bm{\theta}}(\bm{x}) = \prod_{m=1}^{M} P_{\bm{\theta}}(x_m \mid x_1, x_2, \dots, x_{m-1}).
\label{eq:chain-rule}
\end{equation}

As the conditional probabilities are now can be approximated with potential estimation errors, the code length of the compressed symbol $\bm{x}$ is greater than the optimal code length, i.e., the entropy $H(\bm{x})$ defined in (\ref{eq:marginal-entropy}).
In particular, the average code length of the lossless compressed sequence $\bm{x}$ with a probabilistic model $\bm{\theta}$ can be calculated by \cite[Equation 5.23]{mackay2003information}:
\begin{equation}
L(\bm{x}) = H(\bm{x}) + \sum_{x_m \in \mathcal{A}} P(\bm{x}) \log \frac{P(\bm{x})}{P_{\bm{\theta}}(\bm{x})}.
\label{eq:average-code-length}
\end{equation}

In the above equation, the sum $\sum_{x_m \in \bm{x}} P(\bm{x}) \log \frac{P(\bm{x})}{P_{\bm{\theta}}(\bm{x})}$ is a Kullback-Leibler (KL) divergence, which is a non-negative value describing the relative distance between the true probability distribution $P(\bm{x})$ and the estimated probability distribution $P_{\bm{\theta}}(\bm{x})$.
Intuitively, the code length $L(\bm{x})$ (measured in bits) above can be minimized by minimizing the KL divergence, which is also known as the ``free energy" calculated by $\log  \frac{P(\bm{x})}{P_{\bm{\theta}}(\bm{x})}$. In other words, if $P_{\bm{\theta}}(\bm{x}) = P(\bm{x})$, we have $L(\bm{x}) = H(\bm{x})$, which means the probabilistic model with parameters $\bm{\theta}$ can correctly estimate entropy of the data, thus resulting in optimal compression ratio with minimum code length of $H(\bm{x})$ bits.

In our work, we are interested in using a deep learning-based probabilistic model for estimating the marginal probability $P(\bm{x})$ of the voxels. In particular, we utilize a variational autoencoder (VAE) \cite{kingma2013auto} as a core probabilistic model, similar to the approach in \cite{townsend2018practical}. Note that the vanilla model with only fully connected layers in \cite{townsend2018practical} is not applicable for processing 3D data like the voxels in this work. Thus, we develop a complete yet simplistic CVAE model in which 3D convolutional layers are staked on top of the fully connected layers to process the high-dimensional voxel data. The main idea of utilizing the 3D convolutional layers is to subsequently reduce the size of the input vectors before feeding into the fully connected layers. At the same time, the use of these convolutional layers is critical to capture and learn the spatial information of the voxel data.
The model architecture of the Convolutional Variational Autoencoder (CVAE) is illustrated in Fig.~\ref{fig:proposed-method}(b). A detailed explanation of the architecture and the training procedure of the CVAE model can be found in Appendix \ref{appendix}.
 As illustrated in Fig.~\ref{fig:proposed-method}(a), the CVAE model works as a probabilistic model to estimate the true probabilities $P_{\bm{\theta}}(\bm{x}) \approx P(\bm{x})$, with learnable parameters $\bm{\theta}$ (i.e., weights of the deep neural network).  

In the literature, it has been shown that using probabilistic models with deep neural networks \cite{huang2020octsqueeze, nguyen2021learning, wang2021lossy} can produce competitive compression ratios as the deep neural networks can estimate the probability $P(\bm{x})$ with high accuracy. The main idea of the works in \cite{huang2020octsqueeze, nguyen2021learning, wang2021lossy}, and references therein, is that, after learning the probabilistic model from the training data, the learned model is used to produce the probability $P_{\bm{\theta}}(\bm{x})$ for the lossless compression codec, e.g., arithmetic codec or ANS codec. As shown in Fig.~\ref{fig:proposed-method}(c), in conventional methods in \cite{huang2020octsqueeze, nguyen2021learning, wang2021lossy}, the learned probabilistic model $P_{\bm{\theta}}(\bm{x})$ is used as the input of the arithmetic codec to produce the final binary sequence. The main process of a general arithmetic codec or ANS codec is building the cumulative distribution function (CDF) and inverse CDF for the encoder and decoder, respectively, based on the conditional probabilities in equation (\ref{eq:chain-rule}). These CDF and inverse CDF will be used as mapping functions that convert the probability of an input vector to the unit interval between 0 and 1, resulting in a compressed binary representation with expected code length described in equation (\ref{eq:average-code-length}) \cite{witten1987arithmetic, duda2013asymmetric}. 

In this work, we refer to the conventional methods in \cite{huang2020octsqueeze, nguyen2021learning, wang2021lossy} as ``sequential coding" as all the works utilize arithmetic coding with sequential process to compress a batch of point cloud (or voxels) data.  The main idea of the sequential coding process is to iteratively compress the newly arrived input data into an existing compressed sequence, resulting in a final nested message (encoded data). This can be achieved by iteratively retrieving the conditional probabilities $P_{\bm{\theta}}(x_m =i|x_1, \ldots, x_{m-1})$ to produce the final marginal probability $P_{\bm{\theta}}(\bm{x})$, as described in equation (\ref{eq:chain-rule}). When it comes to sequential coding for a batch of $B$ different point clouds, each of which has $M$ voxels, the sequential coding may require $B$ different pairs of CDF-inverse CDF, resulting in $B \times M$ conditional probabilities with large amounts of overhead cost of storage of communication. With this scheme, the potential correlation between the $B$ point clouds, such as shapes and spatial regions of the floating points, is not fully utilized, as the coding/decoding process strictly follows the chain rule to perform compression/decompression steps. 

The overhead cost of requiring $B \times M$ conditional probabilities is usually ignored in the literature works \cite{huang2020octsqueeze, nguyen2021learning, wang2021lossy} as they usually assume the access to these conditional probabilities, or the equivalent CDFs-inverse CDFs, by sharing the pre-trained deep learning models at both encoder (sender) and decoder (receiver) sides. 
However, the use of the pre-trained deep learning model at the receiver may require additional cost of storing or communicating the codec decoder. The cost of this overhead may even exceed the cost of compressing the point cloud data itself, especially in the case of deeper neural networks, or large point cloud datasets. This motivates us to replace the sequential coding method with the novel bits-back coding method.

\subsection{Bits-back Coding for Voxelized Point Cloud Data}
The bits-back coding method was first described by Frey and Hinton in \cite{frey1996free}, and was further developed by Townsend et al. in \cite{townsend2018practical}. The main idea behind bits-back coding is that the encoder (sender) and decoder (receiver) of a given codec, e.g., arithmetic coding \cite{witten1987arithmetic} or asymmetric numeral system (ANS) \cite{duda2013asymmetric}, will share a latent variable model $P(\bm{z})$ (a prior), instead of the marginal probability $P_{\bm{\theta}}(\bm{x})$. The marginal probability $P_{\bm{\theta}}(\bm{x})$ then can be inferred by using Bayes' update rule with the prior $P(\bm{z})$, the posterior $Q_{\bm{\phi}}(\bm{z \mid x})$, and the likelihood $P_{\bm{\theta}}(\bm{x \mid z})$, which will be described later.
The use of a latent variable model is useful to exploit the ``correlation", or ``side information", between the data samples when compressing a batch of such data samples together. The correlation between data samples will gain advantages, in terms of compression ratio, when compressing large batches of data, which is usually the case of compressing data with deep learning-based probabilistic models. 
From the optimal coding rate principle through minimizing the KL divergence in (\ref{eq:average-code-length}), we formalize the bits-back coding approach as follows.

\begin{figure}
\centering
\includegraphics[width=1.0\linewidth]{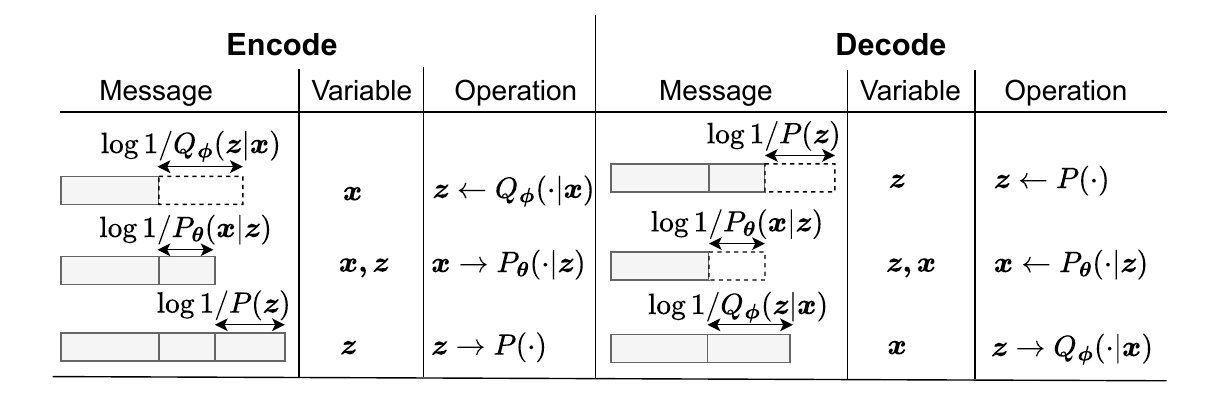}
\caption{Encode (compress) and decode (decompress) process of the bits-back coding. The rectangles with dashed lines denote the decrease in the code length of the message, and rectangles with solid lines denote the increase in the code length. The probability values above the rectangles are linked to the derivative in (\ref{eq:elbo-derivative}). The rectangles are placed in a stack data structure of ANS with the head of the stack on the left-hand side.}
\label{fig:bits-back-ans-operation}
\end{figure}

Starting from equation (\ref{eq:average-code-length}), we now introduce a latent variable $\bm{z}$ which has a multivariate normal (Gaussian) prior $P(\bm{z}) = \mathcal{N}(\bm{z}; 0, \bm{I})$, where $\bm{I}$ is the identity matrix. As we now trying to infer true probability $P(\bm{x})$ from our voxel data with a latent variable model, the marginal probability  $P_{\bm{\theta}}(\bm{x})$ can be written as the total probability of observing $\bm{x}$ under all possible values of the latent variable $\bm{z}$, i.e., 
\begin{equation}
\begin{split}
P_{\bm{\theta}}(\bm{x}) & = \int_{\bm{z}} P_{\bm{\theta}}(\bm{x,z}) d \bm{z}. 
\end{split}
\label{eq:marginal-prob-over-data}
\end{equation}

By taking the $\log$ for the both sides of equation (\ref{eq:marginal-prob-over-data}), introducing the approximate posterior probability $Q_{\bm{\phi}}(\bm{z \mid x})$, and applying Jensen's inequality (i.e., $f\left( \int_{\bm{z}} \bm{z} p(\bm{z}) d\bm{z} \right) \leq \int_{\bm{z}} f(\bm{z}) p(\bm{z}) d\bm{z}$) \cite{kingma2013auto}, we have:
\begin{equation}
\begin{aligned}
&\log P_{\bm{\theta}}(\bm{x}) = \log \int_{\bm{z}} P_{\bm{\theta}}(\bm{x, z}) d \bm{z} \\
&= \log \int_{\bm{z}} Q_{\bm{\phi}}(\bm{z \mid x}) \frac{P_{\bm{\theta}}(\bm{x,z})}{Q_{\bm{\phi}}(\bm{z \mid x})}d \bm{z} \\
&\geq \int_{\bm{z}} Q_{\bm{\phi}}(\bm{z \mid x}) \log \Big( \frac{P_{\bm{\theta}}(\bm{x,z})}{Q_{\bm{\phi}}(\bm{z \mid x})}\Big) d \bm{z} \\
&= \int_{\bm{z}} Q_{\bm{\phi}}(\bm{z \mid x}) \Big[\log P_{\bm{\theta}}(\bm{x \mid z}) \\ & \qquad \qquad \qquad \qquad + \log P(\bm{z}) - \log Q_{\bm{\phi}}(\bm{z \mid x}) \Big] d \bm{z} \\
&= \mathbb{E}_{Q_{\bm{\phi}}(\bm{z \mid x})} \Big[\log P_{\bm{\theta}}(\bm{x \mid z}) + \log P(\bm{z}) - \log Q_{\bm{\phi}}(\bm{z \mid x}) \Big] \\
&= \mathbb{E}_{Q_{\bm{\phi}}(\bm{z \mid x})} \Big[\log P_{\bm{\theta}}(\bm{x \mid z})\Big] - \mathbb{E}_{Q_{\bm{\phi}}(\bm{z \mid x})}\Big[\log \frac{Q_{\bm{\phi}}(\bm{z \mid x})}{P(\bm{z})}\Big]
\end{aligned}
\label{eq:elbo-derivative}
\end{equation}

The lower bound of $\log P_{\bm{\theta}}(\bm{x})$ above is the quantitative value of the ``free energy" of the latent variable modeling process \cite{frey1996free}. Its meaning is two-fold as described as follows. First, in a bits-back encoding process (see Fig.~\ref{fig:bits-back-ans-operation}), this is equivalent to the change of the expected code length \cite{townsend2018practical, Townsend2020HiLLoC} (i.e., line 5 of (\ref{eq:elbo-derivative})):
\begin{equation}
\Delta L = \log P_{\bm{\theta}}(\bm{x \mid z}) + \log P(\bm{z}) - \log Q_{\bm{\phi}}(\bm{z \mid x}).
\label{eq:delta-l}
\end{equation} 
Second, the evidence lower bound (ELBO) value of $\log P_{\bm{\theta}}(\bm{x})$ gives an objective for optimizing the variational autoencoder (VAE) \cite{kingma2013auto} (i.e., line 6 of (\ref{eq:elbo-derivative})):
\begin{equation}
\mathcal{L}_{\bm{\theta;\phi}} = \mathbb{E}_{Q_{\bm{\phi}}(\bm{z \mid x})} \Big[\log P_{\bm{\theta}}(\bm{x \mid z})\Big] - \mathbb{E}_{Q_{\bm{\phi}}(\bm{z \mid x})}\Big[\log \frac{Q_{\bm{\phi}}(\bm{z \mid x})}{P(\bm{z})}\Big].
\label{eq:loss-function}
\end{equation}

Therefore, the main intention of the following methodology in this work is that we first train the CVAE model with training data to obtain the optimized parameters $\{\bm{\theta, \phi}\}$ through minimizing $\mathcal{L}_{\bm{\theta;\phi}}$. After that, we will use the learned CVAE model to encode (compress) and decode (decompress) the voxel data, as illustrated in Fig.~\ref{fig:bits-back-ans-operation}. It can be observed from the derivative in equations (\ref{eq:elbo-derivative}) and (\ref{eq:delta-l}) that, the expected code length of the bits-back coding approach is now can be estimated using the likelihood $ P_{\bm{\theta}}(\bm{x \mid z})$, the prior $P(\bm{z})$, and the approximate posterior probability $Q_{\bm{\phi}}(\bm{z \mid x})$, instead of relying on the explicit chain rule of conditional probabilities as in (\ref{eq:chain-rule}), which can produce significant overhead as described earlier. 

Note that the use of ANS in bits-back coding as an alternative to arithmetic coding in Fig.~\ref{fig:proposed-method} is for compatible implementation purpose only. The main difference between using ANS codec and arithmetic coding codec is that the decoding process of ANS is last-in-first-out (LIFO) as described in Fig.~\ref{fig:bits-back-ans-operation}, while the decoding process of arithmetic coding is first-in-first-out (FIFO) \cite{duda2013asymmetric, townsend2018practical}. There is no advantage in compression ratio when utilizing ANS over arithmetic coding. For more details of bits-back coding with ANS codec, please refer to \cite{townsend2018practical} and \cite{Townsend2020HiLLoC}. 
In the next section, we evaluate the performance of bits-back coding in compressing large point cloud datasets, and show its superior performance in terms of compression ratios and lower overhead cost of storing and communicating the probabilistic decoder. 

\section{Performance Evaluation}
\label{sec:performance-evaluation}
\subsection{Voxelization and Data Processing}
\begin{table}[t]
\centering
\begin{tabular}{|c|c|c|}
\hline
\textbf{Dataset} & \textbf{ShapeNet} & \textbf{Sun RGB-D} \\
\hline
Shape of training set & (8000, 20000, 3) & (10335, 20000, 3) \\
\hline
Shape of test set & (2000, 20000, 3) & (2860, 20000, 3) \\
\hline
Bit-depth ($d$) & 5, 6, 7 & 5, 6, 7 \\
\hline
Data type & float32 & float32 \\
\hline
Data range & [-1.0, 1.0] & [-1.0, 1.0] \\
\hline
\end{tabular}
\caption{Main parameters of the ShapeNet dataset \cite{shapenet2015} and SUN RGB-D dataset \cite{song2015sun} used in this paper.}
\label{tab:datasets}
\end{table}
We first describe two point cloud datasets that we use to validate our bits-back coding approach, followed by the compression process as illustrated in Fig.~\ref{fig:pcc-overview}. The point cloud datasets are the ShapeNet dataset \cite{shapenet2015} and SUN RGB-D dataset \cite{song2015sun}. The ShapeNet dataset contains various 3D object models in different categories such as tables, cars, chairs, and airplanes. The SUN RGB-D dataset contains various raw point cloud data resulting from the reconstruction of RGB-D images, in which each RGB-D image is a pair of an RGB image and a depth image of an indoor scene. We create synthetic point clouds from these above datasets as follows.
From the 3D objects in the ShapeNet dataset, we sample uniformly 20,000 points on the surface of the 3D mesh objects in the dataset. For the RGB-D dataset, we randomly sample 20,000 points from the existing raw point clouds. In other words, each point cloud in our synthetic dataset contains 20,000 points in a 3D coordinator. We use 32 bits of floating-point format (float32) to demonstrate each geometric attribute for each point within the cloud. The final synthetic point cloud dataset is rescaled and normalized within the interval [-1.0, 1.0]. This rescaling and normalization process follows standard practice in common deep learning-based approaches in \cite{nguyen2021learning, wang2021lossy}. The information about the datasets is summarized in Table \ref{tab:datasets}.

After having the synthetic point cloud datasets, we can perform spatial processing with simple voxelization as follows. The interval between -1.0 and 1.0 of each $x, y, z$ axis are equally divided into $2^d$ sub-intervals, resulting in $M = 2^d \times 2^d \times 2^d$ voxels. If there is no point falls inside the region of the voxel, the value of that voxel is zero. Otherwise, the voxel's value is set to 1. By applying the simple voxelization technique above, we can provide ordered input data ready for training our CVAE model. The voxelization technique described above is usually adopted as a processing step before training the entropy model \cite{nguyen2021learning, wang2021lossy} or before building the tree-based compression techniques \cite{huang2020octsqueeze, galligan2018google}. For instance, by applying voxelization with bit-depth value $d=7$ on the ShapeNet's training set described above, we will have a voxelized dataset with the shape (8000, 128, 128, 128) for training our CVAE model. The same data transformation through voxelization applies to the other test sets. We can view the bit-depth parameter as a resolution adjustment parameter with a higher value, i.e., $d=7$ results in a higher resolution voxelized point cloud, and vice versa.

\subsection{Experiment Results}
\subsubsection{Entropy estimation with CVAE model}
\label{subsubsec:data-visualization}

\begin{figure}[t]
\centering
\includegraphics[width=0.95\linewidth]{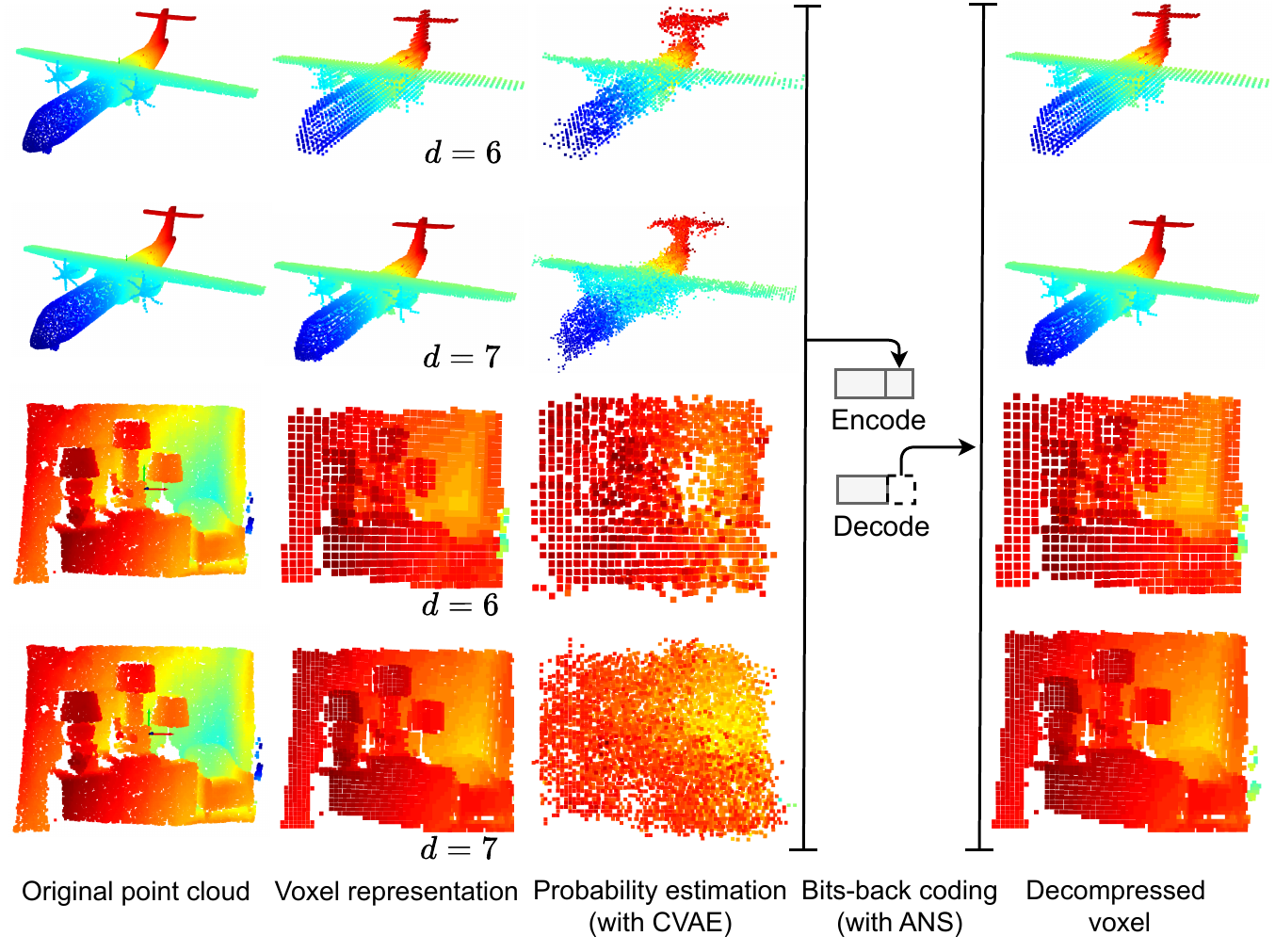}
\caption{Visualization of the original point cloud (first column), voxel representation of the point cloud (second column), reconstructed point cloud from the VAE (third column), and decompressed voxel representation (fifth column). The first and second rows illustrate a data sample from ShapeNet's test set in different bit-depth values, i.e., $d=6$ and $d=7$, respectively. The third and fourth rows illustrate a data sample (a bedroom scene) from the SUN RGB-D's test set with bit-depth values $d=6$ and $d=7$, respectively.}
\label{fig:result-visualization}
\end{figure}

We visualize the main processing steps on point cloud datasets in Fig.~\ref{fig:result-visualization}. For illustration purposes, we take one data sample, i.e., a point cloud of an airplane, from the ShapeNet's test set and one data sample, i.e., a point cloud of a bedroom scene, from the SUN RGB-D's test set. In the second column, we observe that the higher value of bit-depth $d$ results in a better voxel representation of the point cloud. In the third column, we visualize the reconstruction of the voxel data at the output of the CVAE model. The results show that the 3D shape of the objects and scenes are well preserved, but details such as the tail of the airplane and the structure of the bedroom scene are not well recovered. The reason is that the CVAE is optimized based on the variational lower bound of the training data. This means the CVAE will have better generation capabilities, i.e., preserving the 3D shape of the objects and scenes, but lack the fine details of some spatial regions. Note that the blurry voxel reconstruction of the CVAE only affects the compression ratio of the voxel data and since we are using a lossless compression method, the decompressed voxel representation is identical to the original voxel representation, as shown in the third and fifth columns of Fig.~\ref{fig:result-visualization}. This observation suggests that one can potentially develop a more complex entropy estimation model, i.e., building more complex deep neural networks, to gain the 3D spatial details at the third column in Fig.~\ref{fig:result-visualization}, similar to approaches in \cite{nguyen2021learning} and \cite{wang2021lossy}. However, building more complex entropy models is not our main focus and we will let this as a potential research direction, as we can always build better models on top of our framework.

\subsubsection{Impacts of compressing large point cloud dataset}

\begin{figure}[t]
\centering
\includegraphics[width=0.5\linewidth]{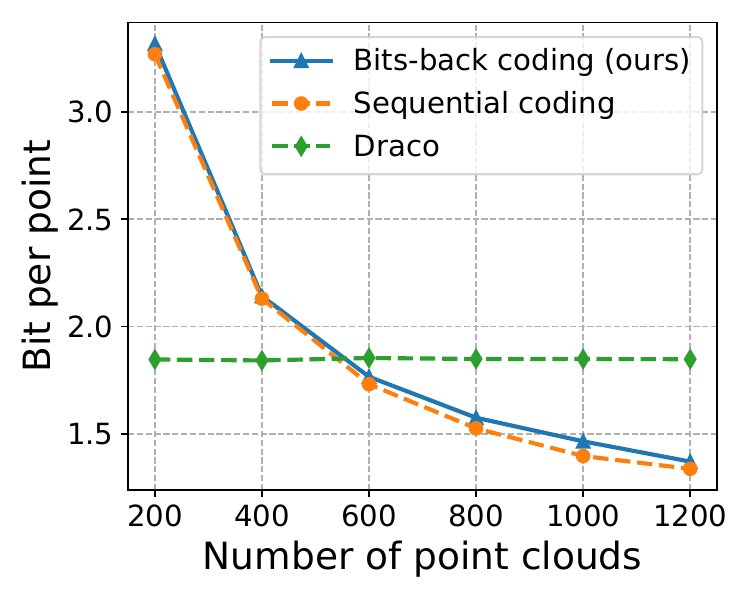}
\caption{Compression ratios (measured in average bit-per-point) of the methods on the Shapenet dataset. The lower the bit-per-point is, the lower the compression ratio (i.e., the ratio of compressed output sequence length to uncompressed input length) can be achieved.}
\label{fig:batch-size-varies}
\end{figure}

Next, we evaluate the performance of the proposed bits-back coding scheme by varying the number of point clouds to be compressed in Fig.~\ref{fig:batch-size-varies}. We use two other approaches that are ``sequential coding" and ``Draco" as baselines for comparison. Sequential coding, as described in Section \ref{subsec:entropy-estimation}, is the technique to sequentially compress the new data sample into an existing compressed message. This technique assumes that both the encoder and decoder can access the approximate probability $P_{\bm{\theta}}(\bm{x})$. This can be done by sharing the deep neural network's parameters for the encoder and decoder. After that, the decoder can sequentially decode the received message by using the pre-trained neural network to produce all conditional probabilities, thus estimating the final marginal probability $P_{\bm{\theta}}(\bm{x})$ using the chain rule in (\ref{eq:chain-rule}) \cite{nguyen2021learning, wang2021lossy}. The second baseline is Draco, a compression library developed by Google that utilizes a tree-based entropy estimation \cite{galligan2018google}. In particular, after the voxelization step, Draco then builds a KD-tree (k-dimensional tree) for removing empty voxels and creating hierarchical regions (nodes), providing a quick way to estimate the entropy of the point cloud across different regions and scales. 

In Fig.~\ref{fig:batch-size-varies}, we show that by increasing the number of point clouds to be compressed, the compression ratios (measured in average bit-per-point) of our bits-back coding approach and the sequential coding approach decrease quickly, meaning that we need a lower number of bits to compress the entire batch of point clouds. For example, by compressing 800 point clouds together, our bits-back coding approach uses 1.56 bits on average to compress a single geometric attribute of the point cloud data. This results in the compressed size of 800 points clouds is 3.1 MB, while the original size of 800 points clouds is 192 MB. 

It can be observed that the sequential coding approach achieves a slightly lower compression ratio (lower bit-per-point value) than ours. The reason is that the sequential coding approach \cite{nguyen2021learning} \cite{wang2021lossy} assumes that the encoder and decoder have direct access to the estimated probabilistic model $P_{\bm{\theta}}(\bm{x})$ on each decoding iteration. This can lead to a near-optimal compression ratio. However, this would lead to a significant amount of memory for storing a batch of individual conditional probabilities for encoding/decoding individual point clouds. We will later show that this approach is not suitable for many practical scenarios as the cost for storage or communication of the codecs may exceed the size of the compressed point clouds. On the other hand, the Draco approach achieves a good compression ratio at 1.8 bit-per-point on average and remains unchanged when the number of point clouds increases. This is because the tree-based approach of Draco for entropy estimation does not allow the sequential coding technique to be applied, meaning that the Draco approach can only compress each point cloud individually. As a result, we can observe that the bits-back coding and sequential coding approach achieve lower compression ratios when compressing more than 600 points cloud together. 

\subsubsection{Impacts of bit-depth}

\begin{figure}[t]
\centering
\begin{subfigure}[t]{0.95\linewidth}
\centering
\includegraphics[width=0.95\linewidth]{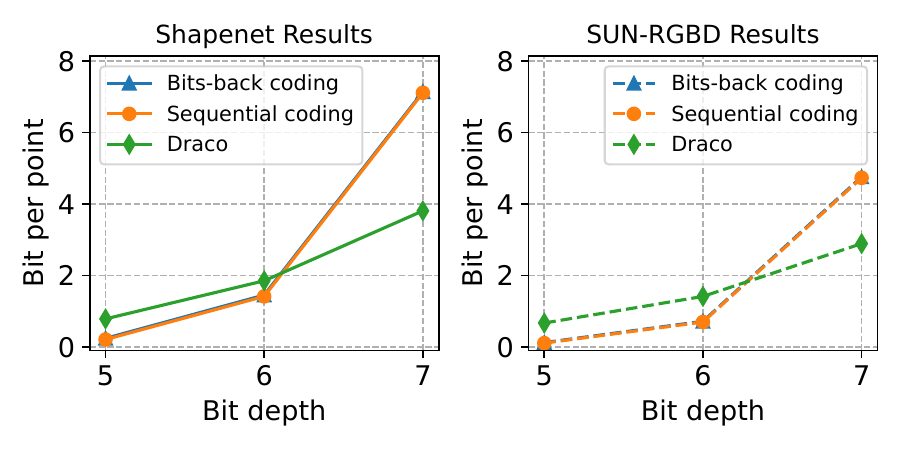}
\caption{}
\end{subfigure}
~
\begin{subfigure}[t]{0.95\linewidth}
\centering
\includegraphics[width=0.95\linewidth]{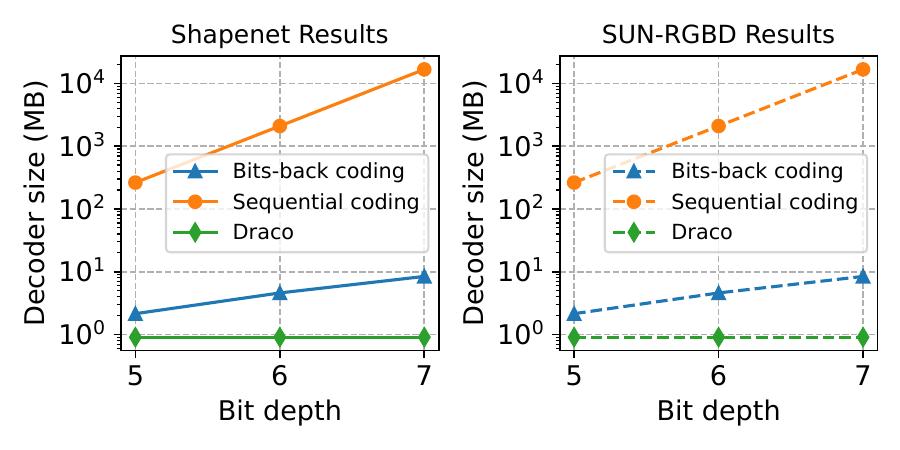}
\caption{}
\end{subfigure}
\caption{(a) Compression ratios of the methods when bit-depth (resolution) increases, and (b) decoder sizes of the methods when bit-depth increases. All the results are obtained when compressing a batch of 1,000 point clouds.}
\label{fig:bit-depth-varies}
\end{figure}

As described above, the sequential coding approach \cite{nguyen2021learning, wang2021lossy} can achieve a near-optimal compression ratio because the encoder and decoder have access to the approximate probability model in each compression iteration. In the following, we will analyze the drawbacks of this approach in practical scenarios and emphasize the advantages of our proposed bits-back coding approach. In Fig.~\ref{fig:bit-depth-varies}, we evaluate the impacts of the bit-depth parameter $d$ on the system performance in terms of compression ratio. 
In particular, as we increase the bit-depth value from $d=5$ to $d=7$, the compression ratio increases as shown in Fig.~\ref{fig:bit-depth-varies}(a). The reason for this trend is the bit-depth parameter indicates the resolution of the voxel representation of the point cloud, as discussed in Section \ref{subsubsec:data-visualization}. As the resolution (i.e., the number of voxels) increases, the compression codec will need to assign a higher number of bits to encode the corresponding point cloud. 

Note that the compression ratio on the ShapeNet dataset (left figure of Fig.~\ref{fig:bit-depth-varies}(a)) is higher than that on the SUN RGB-D dataset (right figure). The difference in compression ratio is caused by the geometric patterns, or the difference in entropy, of the point cloud datasets. We observe that the point clouds in the SUN RGB-D dataset are more scattered around the $x, y, z $ axes in the 3D space, while the point clouds in the ShapeNet dataset are densely distributed around the central region. The spatial patterns suggest that the uncertainty in the estimated entropy $H(\bm{x})$ result in a difference in the average code length, as described in equation (\ref{eq:average-code-length}).

In addition, we observe that the compression ratio's scaling trend of the Draco approach is approximately linear, while the bits-back coding and sequential coding approach exhibit worse scaling with a higher value of bit-depth, i.e., $d=7$. This indicates that the tree-based entropy estimation of Draco is more suitable for higher resolution of voxel representation. Similar findings have been shown in \cite{huang2020octsqueeze}. This suggests that, although our bits-back coding approach achieves a lower compression ratio with lower resolution values compared to Draco, there is still room for improvement. For example, our approach can be extended by incorporating tree-based spatial processing like Octree \cite{huang2020octsqueeze} or KD-tree \cite{galligan2018google} to achieve a lower compression ratio. However, this approach may further complicate the deep neural network architecture and as we stated earlier, this is out of the scope of our main contribution and can be considered as a fruitful future direction. 

Next, we evaluate the decoder sizes (measured in MegaBytes) with respect to the bit-depth value in Fig.~\ref{fig:bit-depth-varies}(b). The decoder sizes of the approaches are calculated by the memory storage required to store the decoder of the codec, which is essential for decoding the compressed data. In the case that the sender and the receiver are separated, e.g., transmitting data over the Internet or a communication channel, the decoder may also need to be exchanged to pre-installed at the receiver, alongside the compressed data. 
As shown in Fig.~\ref{fig:bit-depth-varies}(b), the decoder's size of the sequential coding approach is significantly higher than that of the bits-back coding and Draco approaches.
With the high bit-depth value $d=7$, the overhead of the decoder size of the sequential coding approach is over $10^4$ MB, while the overhead values of the bits-back coding and Draco approaches are less than $10$ MB. The reason is that for compressing a batch of 1,000 point clouds together, the sequential coding approach requires access to the probabilistic model $P_{\bm{\theta}}(\bm{x})$ at each decoding iteration. In other words, the low compression ratio of the sequential coding approach comes from the assumption that the receiver has access to the learned entropy, i.e., $P_{\bm{\theta}}(\bm{x})$ model, at each decoding step, which might not be practical as it requires huge storage and communication overhead. 

The proposed bits-backs coding approach, on the other hand, does not experience the high overhead cost of the decoder size. The bits-back coding does not rely on the assumption of having access to the marginal probability model $P_{\bm{\theta}}(\bm{x})$ at the decoding steps. Instead, the bits-back coding decoder utilizes the prior information $P(\bm{z})$, likelihood $P_{\bm{\theta}}(\bm{x \mid z})$, and posterior $Q_{\bm{\phi}}(\bm{z} \mid \bm{x})$ to decode the compressed data.
Let's take the ``Decode" operation of the bits-back coding approach in Fig.~\ref{fig:bits-back-ans-operation} as the visualized process. 
First, the decoder retrieves the latent variable model $\bm{z}$ by decoding the compressed message with the prior $P(\bm{z})$. Recall that the prior $P(\bm{z})$ is a simple multivariate normal distribution $\mathcal{N}(\bm{z};0, \bm{I})$, where $\bm{I}$ is the identity matrix of size $50 \times 50$, and $\bm{z}$ is the output value of the last layer of the CVAE's encoder, as illustrated in Fig.~\ref{fig:proposed-method}(b). This means that the latent dimension of our CVAE is 50 and is significantly lower than the dimension of the input/output data (i.e., $2^d \times 2^d \times 2^d$). Once the latent variable is retrieved, the message can be further decoded by retrieving the variable $\bm{x}$ through the learned likelihood $P_{\bm{\theta}}(\cdot \mid \bm{z})$.
 Finally, the third step is encoding the variable $\bm{x}$ into the existing message with the learned posterior $Q_{\bm{\phi}}(\cdot \mid \bm{x})$. 
As described in the decoding process above, unlike the sequential coding approach, the decoder of the bits-back coding approach does not need the marginal probability model $P_{\bm{\theta}}(\bm{x})$, thus resulting in lower overhead cost caused by accessing the marginal probability model at each decoding iteration. By utilizing the lower-dimensional latent variable, along with the posterior during the encoding-decoding process, bits-back coding enables efficient decoding while effectively handling correlations between data samples, especially when compressing large batches of point clouds.

The lowest overhead cost of the Draco approach is reflected by its tree-based spatial processing. In particular, the decoder of the Draco approach may require only the information of the tree data structure such as the head of the KD-tree and the depth of the tree, to fully recover the tree data structure for later decoding steps. For this, the decoding can be done by tracing the paths through the tree to recover the geometric attributes of the point clouds. However, the lightweight capability of Draco's decoder results in an inefficient compression ratio when encoding a large batch of point clouds as shown in Fig.~\ref{fig:batch-size-varies}. Notably, our proposed bits-back coding approach might be potentially combined with tree-based spatial processing with an ANS codec like Draco to achieve better compression ratios at high bit-depth values. This may require a complex entropy estimation model, e.g., using multiple deep neural networks together, for encoding and decoding serialized tree data \cite{huang2020octsqueeze}. 

\section{Conclusion}
In this work, we have proposed a novel bits-back coding method for compressing large point cloud datasets. The novelty of the approach is based on the bits-back coding method, which utilizes the latent variable model over the model parameters to encode and decode point cloud data. By utilizing the prior, likelihood, and approximate posterior of the CVAE model during the decoding step, our approach achieves a competitive compression ratio compared to the sequential coding approach with deep neural networks. The use of the bits-back coding method enables us to achieve a comparable compression ratio to the other sequential coding with deep learning models while consuming  lower memory and communication costs. Notably, the bits-back coding approach does not require access to the learned entropy model $P_{\bm{\theta}}(\bm{x})$ during the decoding process, thus enabling lightweight and practical decoder implementation. Future research directions could be adopting advanced spatial processing techniques, such as building tree-based models, using posterior estimation with flow-based models, and building deeper neural networks  for better entropy estimation.

\vspace{-0.2cm}
\appendix
\section{Architecture of the Proposed CVAE}
\label{appendix}
In particular, the CVAE's encoder consists of three Convo3D (3D convolution) layers, followed by two fully-connected layers. The main idea of using stacked Convo3D layers is to subsequently reduce the dimensions of the input data before feeding it into the fully connected layers. For high-dimensional data like our voxels, where the bit-depth value $d=7$ results in dimensions of $M=2^7 \times 2^7 \times 2^7$, the input data expands to a flattened array with $2^{21}$ features (more than 2 million). Processing such a large flattened array using a vanilla VAE with only fully-connected layers becomes impractical due to the immense computational demand. Instead, by using the Convo3D layers, the input dimension is reduced from $M$ to 2,000 features before forwarding these 2,000 features into the fully connected layers. The parameters of each Convo3D and fully-connected layer are illustrated in detail in Fig.~\ref{fig:proposed-method}(b). In this work, our implementation of different bit-depth values with $d=5$, $d=6$, and $d=7$, with only simple modifications of the kernel size and stride parameters of the Convo3D layers. For higher bit-depth (i.e., $d > 7$), it is possible to further implement such bigger and deeper models. In return, it would require large computing power and running time while does not provide further insights from our work, which extensively focuses on investigating the potential of the bits-back coding. 

After passing the $M$ voxel features through the CVAE's encoder, the number of features is reduced to 500, which is used for constructing a latent space model that consists of 50 features. The addition module before the CVAE's latent space in Fig.~\ref{fig:proposed-method}(b) is to illustrate the reparameterization trick, in which the latent space is subjected to a Gaussian prior, i.e., $P(\bm{z}) \sim \mathcal{N}(\bm{z};0, \mathbf{I})$ \cite{kingma2013auto}.
Following the latent space layer are two fully connected layers with the same sizes as the ones of the CVAE's encoder. Finally, the last three layers are the ConvoTran3D (3D transposed convolution) layers. The ConvoTran3D layers are utilized to subsequently increase the dimensional features of the input vector, so that the final output of the CVAE model has the same dimension, i.e., $M$, as the input vector.
We then train the CVAE for the two datasets described in Table \ref{tab:datasets}. The training process is done by using Adam optimizer with a learning rate of 0.001. The loss function is conventional ELBO in (\ref{eq:loss-function}). The activation function used across the hidden layers is ReLu. The final output layer's activation function is Sigmoid as we are modeling the binarized voxel data. The Sigmoid output of the CVAE model then can be used as a parametric Bernoulli distribution to produce the voxels having values of either 0 or 1. We train the CVAE model for 500 training epochs. After training, the CVAE model can produce an estimated probability for the input voxel as shown in Fig.~\ref{fig:result-visualization}. Our details of implementation and reproducibility can be found at \text{https://github.com/hieunq95/gpcc-bits-back}.

\vspace{-0.3cm}
\bibliographystyle{IEEEtran}
\bibliography{references}

\begin{thebibliography}{10}
\providecommand{\url}[1]{#1}
\csname url@samestyle\endcsname
\providecommand{\newblock}{\relax}
\providecommand{\bibinfo}[2]{#2}
\providecommand{\BIBentrySTDinterwordspacing}{\spaceskip=0pt\relax}
\providecommand{\BIBentryALTinterwordstretchfactor}{4}
\providecommand{\BIBentryALTinterwordspacing}{\spaceskip=\fontdimen2\font plus
\BIBentryALTinterwordstretchfactor\fontdimen3\font minus
  \fontdimen4\font\relax}
\providecommand{\BIBforeignlanguage}[2]{{%
\expandafter\ifx\csname l@#1\endcsname\relax
\typeout{** WARNING: IEEEtran.bst: No hyphenation pattern has been}%
\typeout{** loaded for the language `#1'. Using the pattern for}%
\typeout{** the default language instead.}%
\else
\language=\csname l@#1\endcsname
\fi
#2}}
\providecommand{\BIBdecl}{\relax}
\BIBdecl

\bibitem{cao20193d}
C.~Cao, M.~Preda, and T.~Zaharia, ``3d point cloud compression: A survey,'' in
  \emph{Proceedings of the 24th International Conference on 3D Web Technology},
  Jul. 2019, pp. 1--9.

\bibitem{huang2020octsqueeze}
L.~Huang, S.~Wang, K.~Wong, J.~Liu, and R.~Urtasun, ``Octsqueeze:
  Octree-structured entropy model for lidar compression,'' in \emph{Proceedings
  of the IEEE/CVF Conference on Computer Vision and Pattern Recognition}, Jun.
  2020, pp. 1313--1323.

\bibitem{nguyen2021learning}
D.~T. Nguyen, M.~Quach, G.~Valenzise, and P.~Duhamel, ``Learning-based lossless
  compression of 3d point cloud geometry,'' in \emph{IEEE International
  Conference on Acoustics, Speech and Signal Processing}.\hskip 1em plus 0.5em
  minus 0.4em\relax IEEE, Jun. 2021, pp. 4220--4224.

\bibitem{wang2021lossy}
J.~Wang, H.~Zhu, H.~Liu, and Z.~Ma, ``Lossy point cloud geometry compression
  via end-to-end learning,'' \emph{IEEE Transactions on Circuits and Systems
  for Video Technology}, vol.~31, no.~12, pp. 4909--4923, Jun. 2021.

\bibitem{he2022density}
Y.~He, X.~Ren, D.~Tang, Y.~Zhang, X.~Xue, and Y.~Fu, ``Density-preserving deep
  point cloud compression,'' in \emph{Proceedings of the IEEE/CVF Conference on
  Computer Vision and Pattern Recognition}, 2022, pp. 2333--2342.

\bibitem{witten1987arithmetic}
I.~H. Witten, R.~M. Neal, and J.~G. Cleary, ``Arithmetic coding for data
  compression,'' \emph{Communications of the ACM}, vol.~30, no.~6, pp.
  520--540, 1987.

\bibitem{duda2013asymmetric}
J.~Duda, ``Asymmetric numeral systems: Entropy coding combining speed of
  huffman coding with compression rate of arithmetic coding,'' \emph{arXiv
  preprint arXiv:1311.2540}, 2013.

\bibitem{mackay2003information}
D.~J. MacKay, \emph{Information Theory, Inference and Learning
  Algorithms}.\hskip 1em plus 0.5em minus 0.4em\relax Cambridge university
  press, 2003.

\bibitem{frey1996free}
B.~J. Frey and G.~E. Hinton, ``Free energy coding,'' in \emph{Proceedings of
  Data Compression Conference}.\hskip 1em plus 0.5em minus 0.4em\relax IEEE,
  Mar. 1996, pp. 73--81.

\bibitem{townsend2018practical}
J.~Townsend, T.~Bird, and D.~Barber, ``Practical lossless compression with
  latent variables using bits back coding,'' in \emph{International Conference
  on Learning Representations}, May 2019.

\bibitem{Townsend2020HiLLoC}
J.~Townsend, T.~Bird, J.~Kunze, and D.~Barber, ``Hilloc: lossless image
  compression with hierarchical latent variable models,'' in
  \emph{International Conference on Learning Representations}, Apr. 2020.

\bibitem{galligan2018google}
\BIBentryALTinterwordspacing
F.~Galligan, M.~Hemmer, O.~Stava, F.~Zhang, and J.~Brettle, ``Google/draco: A
  library for compressing and decompressing 3d geometric meshes and point
  clouds,'' 2018. [Online]. Available: \url{https://github.com/google/draco}
\BIBentrySTDinterwordspacing

\bibitem{kingma2013auto}
D.~P. Kingma and M.~Welling, ``Auto-encoding variational bayes,''
  \emph{International Conference on Learning Representations}, Apr. 2014.

\bibitem{shapenet2015}
A.~X. Chang, T.~Funkhouser, L.~Guibas, P.~Hanrahan, Q.~Huang, Z.~Li,
  S.~Savarese, M.~Savva, S.~Song, H.~Su, J.~Xiao, L.~Yi, and F.~Yu,
  ``{ShapeNet: An information-rich 3D model repository},'' Stanford University
  --- Princeton University --- Toyota Technological Institute at Chicago, Tech.
  Rep. arXiv:1512.03012 [cs.GR], 2015.

\bibitem{song2015sun}
S.~Song, S.~P. Lichtenberg, and J.~Xiao, ``Sun rgb-d: A rgb-d scene
  understanding benchmark suite,'' in \emph{Proceedings of the IEEE Conference
  on Computer Vision and Pattern Recognition}, Jun. 2015, pp. 567--576.

\end{thebibliography}

\end{document}